\pdfoutput=1

\documentclass[11pt]{article}

\usepackage[final]{acl}

\usepackage{array}
\newcolumntype{R}[1]{>{\raggedleft\hspace{0pt}}m{#1}}

\usepackage{amsmath}

\usepackage{times}
\usepackage{latexsym}

\usepackage[T1]{fontenc}

\usepackage[utf8]{inputenc}

\usepackage{microtype}

\usepackage{inconsolata}
\usepackage{todonotes}

\usepackage{graphicx}

\usepackage{authblk}

\usepackage{amssymb}
\title{Deep-change at AXOLOTL-24: Orchestrating WSD and WSI Models for Semantic Change Modeling}
\author[1,2]{Denis Kokosinskii}
\author[1,3]{Mikhail Kuklin}
\author[4]{Nikolay Arefyev}
\affil[1]{Moscow State University, Russia}
\affil[2]{SaluteDevices, Russia}
\affil[3]{Yandex, Russia}
\affil[4]{University of Oslo, Norway}
\affil[ ]{\texttt{kokosinskiidv@my.msu.ru, kuklin.mike@yandex.ru, nikolare@uio.no}}

\begin{document}
\maketitle

\begin{abstract}
This paper describes our solution of the first subtask from the AXOLOTL-24 shared task on Semantic Change Modeling. The goal of this subtask is to distribute a given set of usages of a polysemous word from a newer time period between senses of this word from an older time period and clusters representing gained senses of this word. We propose and experiment with three new methods solving this task. Our methods achieve SOTA results according to both official metrics of the first substask. Additionally, we develop a model that can tell if a given word usage is not described by any of the provided sense definitions. This model serves as a component in one of our methods, but can potentially be useful on its own.

\end{abstract}

\section{Introduction}
\begin{figure}[t]
    \centering
    \includegraphics[clip,width=\linewidth]{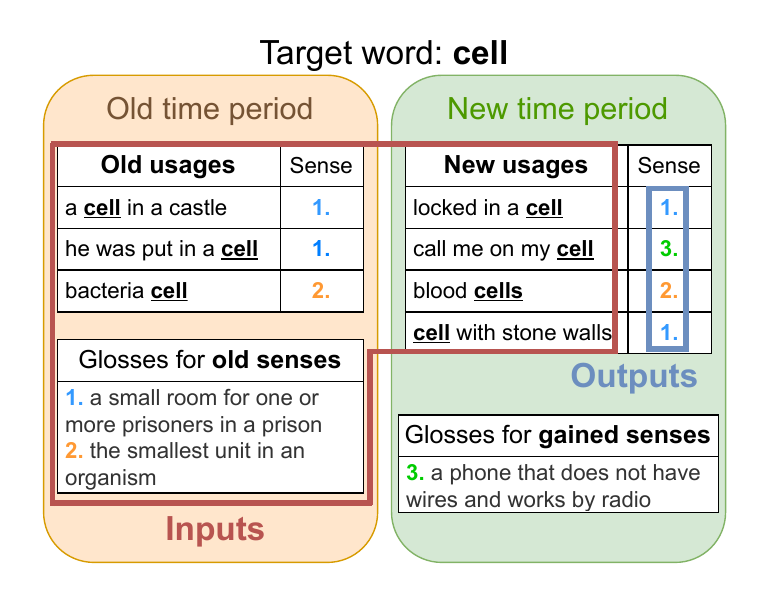}
    \caption{An example of data for the first subtask of AXOLOTL-24.}
    \label{fig:terms}
\end{figure}

The shared task on explainable Semantic Change Modeling (SCM) AXOLOTL-24~\citep{fedorova-etal-2024-axolotl} is related to automation of Lexical Semantic Change (LSC) studies, i.e. linguistic studies on how word meanings change over time. It consists of two subtasks, however, we focus on the first one and skip the definition generation subtask.
Unlike other shared tasks LSC held before, the first subtask of AXOLOTL-24 requires automatic annotation of individual usages of target words instead of target words as a whole. An example of the provided data and required outputs is shown on Figure~\ref{fig:terms}. Namely, for each target word, two sets of usages from an older and a newer period are given (we will call them \textit{old} and \textit{new} usages). Additionally, a set of glosses describing word senses in the older time period (\textit{old senses}) are provided, and the old usages are annotated with these sense glosses. Senses occurring among the new usages (\textit{new senses}) should be discovered automatically. To be precise, the goal is to annotate each new usage with one of the given old sense glosses, or a unique sense identifier if none of them is applicable. We will refer to those senses that occur only among old and only among new usages as \textit{lost} and \textit{gained} senses, and all other senses as \textit{stable} senses.

To solve the task, we experiment with three types of models. Word Sense Disambiguation (WSD) models for a given word usage select among given glosses the most suitable one. Word Sense Induction (WSI) models group word usages into clusters corresponding to word senses, they are applicable even when sense descriptions are not available. Finally, Novel Sense Detection (NSD) models find usages corresponding to unknown word senses, the ones that are not covered by the provided definitions. We propose three methods that solve the task. Our best solution denoted as Outlier2Cluster combines all three types of models in a novel way, essentially using an NSD model for each usage to decide whether to return a definition selected by a WSD model, or an identifier of a cluster this usage was put into by a WSI model. On average across languages, this solution achieves SOTA results among all participants of the first substask of AXOLOTL-24 according to both official metrics.

An important additional contribution is the proposed NSD model and the related experiments. We study the importance of different features of the NSD model and its effect on SCM quality. Our experiments suggest that improving NSD quality is the most promising direction for the future.

\section{Related work}
\textbf{LSCD methods.}  Several shared tasks related to LSCD were organized in the past, including~\citet{schlechtweg-etal-2020-semeval, rushifteval2021, Zamora2022lscd}. Unlike AXOLOTL-24~\citep{fedorova-etal-2024-axolotl}, they required word-level predictions from their participants, either in the form of word ranking or binary word classification. This type of task setup is generally mentioned under the name Lexical Semantic Change Detection / Discovery (LSCD). In the earlier shared tasks the best results were achieved by solutions that employed non-contextualized word-level embeddings such as word2vec~\citep{NIPS2013_9aa42b31} and vector alignment methods such as Canonical Correlation Analysis and Orthogonal Procrustes
Alignment~\citep{pomsl-lyapin-2020-circe,prazak-etal-2020-uwb}. However, recently token-level methods~\citep{laicher-etal-2021-explaining, rachinskiy-arefyev-2021-glossreader,rachinskiy-arefyev-2022-black} have surpassed them. These methods rely on masked language models fine-tuned on existing datasets for various tasks of lexical semantics. For instance, solutions relying on the contextualized embeddings from GlossReader, which is a WSD system, have shown SOTA results in the shared tasks on LSCD in Russian and Spanish~\citep{rachinskiy-arefyev-2021-glossreader,rachinskiy-arefyev-2022-black}. Methods proposed in this work exploit GlossReader too, both as a WSD model and as a source of contextualized embeddings well-suited for LSC-related tasks. 

GlossReader is a multilingual gloss-based WSD model originally developed to solve the Word-in-Context task~\citep{rachinskiy-arefyev-2021-glossreader-wic}. It modifies the English WSD model BEM~\citep{blevins-zettlemoyer-2020-moving} replacing the backbone with the multilingual XLM-R language model~\citep{conneau2019unsupervised}. The model consists of a gloss encoder and a context encoder, both initialized with the XLM-R weights and fine-tuned jointly learning to select among all glosses of a target word the one describing its sense in a given context. Specifically, the dot product between the context embedding and the correct gloss embedding is maximized. 

\textbf{NSD methods.} Several methods were proposed to solve the NSD task. Some of them perform WSI internally. 
For instance, \citet{lau-etal-2012-word, cook-etal-2014-novel} employ a topic modelling approach to jointly cluster old and new usages using the Hierarchical Dirichlet Process. Clusters are ranked based on the novelty score (the difference between estimated probabilities of a cluster appearing in the new and the old corpus). While the method was originally designed for LSCD, the novelty ranking of senses can be combined with a static threshold to identify novel senses. 

Alternatively, \citet{MITRA_MITRA_MAITY_RIEDL_BIEMANN_GOYAL_MUKHERJEE_2015} performs WSI separately for an old and a new corpus on graphs, where an edge weight between two words is proportional to the number of words appearing in bigrams with both of them. A cluster in the new corpus is labeled as a novel sense if words in this cluster have weak links with the target word in the graph for the old corpus. 
A recent method by~\citet{ma-etal-2024-graph} uses BERT \citep{devlin-etal-2019-bert} to build contextualized representations. It employs agglomerative clustering to perform WSI and then matches old and new clusters based on their centroids. The new clusters that are not matched are considered novel senses. Similarly to this method we use agglomerative clustering for WSI, but employing GlossReader to obtain contextualized embeddings.

In~\citet{erk-2006-unknown} several NSD methods were proposed to detect word senses that are not described in FrameNet \citep{baker-etal-1998-berkeley-framenet}. Instead of relying on WSI, similarly to our NSD method their best method formulates the task as an outlier detection problem. They employ distances between old and new usages requiring a significant number of old usages for each sense, which are not always available in AXOLOTL-24. Thus, we rely on distances between new usages and old glosses instead. Another similar method is introduced in~\citet{lautenschlager2024detectionnonrecordedwordsenses}. They use the XL-LEXEME model \citep{cassotti-etal-2023-xl} to build representations for usages and senses. Sense representations are built from glosses or example usages of senses taken from dictionaries. They do not always contain the target word, which makes application of XL-LEXEME non-trivial. Authors attempt to solve this problem by modifying glosses and example usages to include the target word. For each usage its nearest sense is found based on the cosine similarity or the Spearman's correlation between their embeddings. If the similarity is above a threshold, the usage is considered to belong to some non-described sense. 
Our methods also rely on usage and sense representations, but we use GlossReader which has a separate gloss encoder and does not require any preprocessing for glosses. We experiment with many measures of similarity between a sense and a usage embedding, and found the manhattan distance between l1-normalized embeddings to outperform other measures and a classifier on a combination of measures to perform best. However, we did not experiment with example usages from sense inventories. When such usages are available, being combined with glosses they may potentially improve sense representations.


\section{Methods}
\subsection{Target word positions}
All our methods assume that a usage is represented as a string and two character-level indices pointing to a target word occurrence inside this string. However, for the Russian subsets these indices were absent. To find them, we first generated all grammatical forms for each target lemma using Pymorphy2~\citep{pymorphy2}. Then retrieved all occurrences of these forms as separate tokens in the provided usages employing regular expressions.\footnote{E.g. \texttt{'\textbackslash b(cat|cats)\textbackslash b'}, where \texttt{\textbackslash b} denotes a word boundary. Matching is case-insensitive.}
For usages with several occurrences of the target word we selected one of them that has both left and right context of reasonable length.\footnote{This idea is based on our observations that a word occurrence is encoded sub-optimally when it is either the first or the last token, which is probably related to confusion of Transformer heads that have learnt to attend to the adjacent tokens~\cite{voita-etal-2019-analyzing}. The heuristic implemented takes the second to last occurrence if there are more than two of them. For two occurrences it takes $argmax_{u \in \{u1,u2\}} min(l_u,r_u)$, where $l_u,r_u$ are the lengths of the left and the right contexts.} 
We inspected new usages from the development and the test sets that did not contain any of the automatically generated word forms and added absent forms manually, then reran retrieval.\footnote{Repeating this manual procedure for all Russian data requires significantly more efforts and would have few benefits for our methods. Thus, all old usages having this issue were left without indices and new usages from the training set we dropped.}

\subsection{WSD methods}
The first group of methods in our experiments include pure WSD methods, which select one of the provided definitions of old senses for each new usage, and thus, cannot discover gained senses.

\textbf{GlossReader.} We employ the original GlossReader model~\citep{rachinskiy-arefyev-2021-glossreader-wic} as the baseline. For a given \textbf{new} usage $u$ of a target word $w$ its usage representation $r_{u}$ is built with the context encoder. Then gloss representations $r_{g}$ are built for each gloss $g$ of the target word $w$ using the gloss encoder. Finally, the gloss with the highest dot product similarity to the usage is selected. 

To improve the results, we further fine-tune the GlossReader model on the data of AXOLOTL-24.

\textbf{GlossReader FiEnRu} is fine-tuned following the original GlossReader training procedure on three datasets: the train sets of the shared task in Finnish and Russian, and the English WSD dataset SemCor \citep{SemCor} which GlossReader was originally trained on. We employ all old and new usages from the Russian and Finnish datasets along with their sense definitions. We fined-tuned for 3 epochs using 90/10\% train/validation split to select the best checkpoint.\footnote{The last checkpoint was selected, though after $\approx$0.5 epochs metrics improve very slowly.}

\textbf{GlossReader Ru} is fine-tuned exactly the same way, but only on the train set in Russian.

\textbf{GlossReader Fi SG} is fine-tuned on the Finnish train set only. Unlike two previous models, we made an attempt to teach this model how to discover novel senses. Specifically, we replaced all glosses of gained senses with a Special Gloss (SG) "the sense of the word is unknown" in Finnish\footnote{"sanan merkitystä ei tunneta" as translated by Google Translate} and fine-tuned the model as before. For inference we tried adding the special gloss to the provided old glosses, essentially extending the WSD model with NSD abilities. However, this resulted in a noticeable decrease of the metrics on the Finnish development set. Thus, we decided to use the special gloss for training only.\footnote{The majority of words in the Finnish dataset have one sense only, see Section~\ref{sec:data stats}. Pure WSD methods always return perfect predictions for such cases, thus, it is very hard to compete with them on this dataset. In the future we plan to experiment with this model on the Russian dataset having much smaller proportion of such words.}

\subsection{WSI methods}
Unlike WSD methods, WSI methods do not use definitions or any other descriptions of word senses. Instead they discover senses of a word from an unlabeled set of its usages by splitting this set into clusters hopefully corresponding to word senses. WSI methods cannot attribute usages to the provided old glosses, but can potentially group usages of the same sense, including gained senses, into a separate cluster.

\textbf{Agglomerative} is the only WSI method we propose and experiment with. For each new usage its representation $r_{u}$ is built using the context encoder of the original GlossReader model. Then we perform agglomerative clustering of old usages using the cosine distance and average linkage on these representations. This clustering algorithm was successfully used to cluster vectors of lexical substitutes, another kind of word sense representations, in several substitution-based WSI methods~\cite{amrami2018word, amrami2019towards, arefyev2020always, kokosinskii2024multilingual}, as well as for LSCD \cite{laicher-etal-2021-explaining, ma-etal-2024-graph}.

Agglomerative clustering starts with each usage in a separate cluster, then iteratively merges two closest clusters. The distance between two clusters is the average pairwise cosine distance from the usages in the first cluster to the usages in the second one. Merging stops when the predefined number of clusters is reached. We range the number of clusters between 2 and 9 and select a clustering with the highest Calinski-Harabasz score \citep{calinski1974dendrite}.\footnote{For one or two usages the Calinski-Harabasz score is not defined. We return a single cluster in such cases.}

\subsection{SCM methods}
WSD and WSI methods provide only partial solutions of the semantic change modeling task, the former cannot discover novel senses, and the latter cannot annotate usages with the old glosses provided. We propose three new methods developed to fully solve the task.

\subsubsection{AggloM}
Our first SCM method modifies the Agglomerative WSI method by incorporating old usages and senses into the clustering process. We perform agglomerative clustering of a set containing both old and new usages of a target word. Initially, each new usage is assigned to a separate cluster. The old usages are clustered according to the provided sense annotations. Then at each iteration we compute the distances from each cluster containing only new usages to all other clusters. The distance between two clusters is defined as the minimum cosine distance between the usage representations from the first and the second cluster.\footnote{This is known as single linkage.} We then merge two nearest clusters, one of which contains new usages only. This iterative merging process stops when the number of clusters is larger than the number of old senses by $k \geq 0$. Therefore AggloM returns exactly $k$ novel senses, where $k$ is a hyperparameter.\footnote{In the preliminary experiments on the Finnish development set we selected $k=0$, which means that all new usages are eventually merged into clusters representing old senses. This is likely related to the low proportion of gained senses in this dataset and noisy usages which make them hard to discover.} We do not use this method on the Russian datasets because for most senses there are no old usages there.

\textbf{AggloM FiEnRu} is identical to AggloM but relies on the fine-tuned GlossReader FiEnRu.

\subsubsection{Cluster2sense} In the second SCM method we first independently cluster new usages using the Agglomerative WSI method and annotate them with the old senses using GlossReader FiEnRu. We then keep the clustering obtained from WSI, but relabel those clusters that overlap heavily with one of the predicted senses. Specifically, we label a cluster $c$ with a sense $s$ if $c$ has the highest Jaccard similarity to $s$ among all the old senses of the target word, and at the same time $s$ has the highest similarity to $c$ among all the clusters built for new usages of this word. Notably, two clusters cannot be labeled with a single sense, thus the clustering of usages is identical to the one originally predicted by WSI. Some clusters will not be labeled with any sense, thus, Cluster2sense can discover gained senses. At the same time, some senses will not be assigned to any cluster, which means the potential to discover lost senses as well.

\begin{figure}
    \centering
    \includegraphics[trim={54 655 309 40},clip,width=\linewidth]{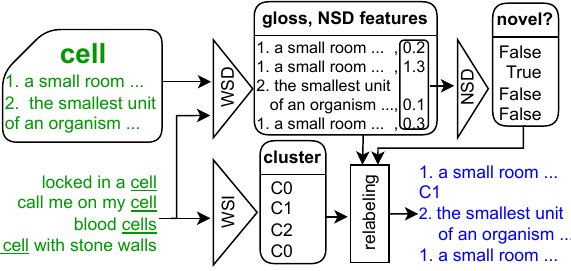}
    \caption{Outlier2Cluster pipeline. Inputs are in green and outputs are in blue. Triangles denote ML models.}
    \label{fig:outlier2sense-diagram}
\end{figure}

\subsubsection{Outlier2Cluster}
\label{sec: outlier2cluster}
Unlike Cluster2Sense which relabels whole clusters, Outlier2Cluster relabels individual usages. Figure~\ref{fig:outlier2sense-diagram} shows the processing pipeline. First WSD and WSI predictions are independently made by GlossReader FiEnRu and Agglomerative respectively. 
Then we discover usages of gained senses. For that we propose a Novel Sense Detection (NSD) model finding usages of those senses that we do not have definitions for.\footnote{In the context of the shared task these are gained senses. However, the approach is general enough to discover lost senses when a modern dictionary and old usages are given, or just senses from the same time period as the dictionary but not covered by it.}
Finally, we return WSI predictions for all these discovered usages, and WSD predictions for all other usages.

\textbf{Novel sense detection.} We treat the NSD task as an outlier detection problem, essentially finding those usages that are distant enough from all the provided definitions. Since GlossReader selects the most similar definition for a given usage, it is enough to check if this definition is distant enough to conclude that the usage is an outlier. To check this we employ a logistic regression classifier. Each input example corresponds to a single usage and a gloss selected for this usage by the WSD model. The output is 1 if this usage is an outlier, i.e. does not belong to the predicted sense, and 0 otherwise. 

We use distances (computed with several distance functions) between GlossReader representations of the new usages and the glosses for old senses as features for logistic regression. 

For the new usage $u$ and the selected definition $g$ we take the corresponding representations $r_{u}$ and $r_{g}$ from a gloss encoder and a context encoder respectively. We take these representations from two different GlossReader models, the original one and GlossReader FiEnRu, and calculate distances from $r_{u}$ to $r_{g}$ using different distance and normalization functions. This gives 10 different features presented in Table \ref{tab:features}. We also include three extra features: the number of old usages, old senses, and new usages for the target word in the dataset. We employ the Standard Scaler to normalize features and train the logistic regression with L2 regularization of $C=1$. 

\begin{table}
    \centering
    \small
    \begin{tabular}{|l|l|lll|}
         \hline
         \multicolumn{2}{|r|}{\textbf{Distance Function}} & \textbf{Cos.} &\textbf{Euclid.} & \textbf{Manh.} \\
         \hline
         \textbf{Encoders} & \textbf{Normalized} & & & \\
         GR FiEnRu & No & \checkmark & \checkmark & \checkmark \\
         GR FiEnRu & L1-norm &  &  & \checkmark \\
         GR FiEnRu & L2-norm &  & \checkmark & \\
         GR & No & \checkmark & \checkmark & \checkmark \\
         GR & L1-norm & & & \checkmark \\
         GR & L2-norm & & \checkmark & \\
         \hline
    \end{tabular}
    \caption{Ten distance-based features used in the NSD model. Distances are calculated between usage and gloss representations obtained from context and gloss encoders of the same GlossReader model. GR stand for GlossReader, Cos. is the cosine distance, Euclid. is the euclidean distance, Manh. is the manhattan distance. }
    \label{tab:features}
\end{table}

Thus, the trained logistic regression can be used for each usage to decide whether the WSD method has assigned a correct sense or should be replaced with some cluster corresponding to a gained sense. If the score is above a threshold of 0.65, which was selected on the development sets of the shared task, the usage is considered an outlier.

\begin{table*}
\begin{center}

\footnotesize
\begin{tabular}{|lllllll|}
\hline
& &\textbf{Requires} &\textbf{Requires}&\textbf{Requires a}  &\textbf{Able to }& \textbf{Able to}  \\
& \textbf{Underlying} &  \textbf{usages of}   & \textbf{old} & \textbf{train set with}& \textbf{discover} & \textbf{predict} \\
& \textbf{embeddings} & \textbf{old senses} & \textbf{glosses} & \textbf{gained senses$^*$} & \textbf{gained senses} & \textbf{old senses} \\
\hline
GR &GR &- &\checkmark &- &- &\checkmark \\
GR FiEnRu &GR FiEnRu &- &\checkmark &- &- &\checkmark \\
GR Ru &GR Ru &- &\checkmark &- &- &\checkmark \\
GR Fi SG &GR Fi SG &- &\checkmark &\checkmark &- &\checkmark \\
\hline
Agglomerative &GR &- &- &- &\checkmark &- \\
\hline
AggloM &GR &\checkmark &- &- &if $k>0$ &\checkmark \\
AggloM FiEnRu &GR FiEnRu &\checkmark &- &- &if $k>0$ &\checkmark \\
Cluster2Sense &GR, GR FiEnRu &- &\checkmark &- &\checkmark &\checkmark \\
Outlier2Cluster &GR, GR FiEnRu &- &\checkmark &\checkmark &\checkmark &\checkmark \\
\hline
\end{tabular}    
\end{center}
\caption{A brief description of the proposed methods. GR stands for GlossReader model. $^*$GR Fi SG is trained to predict the special gloss for usages of all gained senses. In Outlier2Cluster the NSD model is trained to detect usages of gained sense. }
\label{tab: methods}
\end{table*}

We train two NSD models on the Russian and the Finnish development sets separately and use the trained models for the corresponding test sets. For the surprise language, we do not have labeled data to select one of two models or train a separate model, thus, we simply report the results of both models.

\textbf{Outlier relabeling.} We experiment with two ways of assigning clusters to the detected outliers. Our first approach (\textbf{\mbox{w/o WSI}}) groups all outliers into a single new cluster. Alternatively, \textbf{\mbox{w/ WSI}} approach assigns the clusters predicted by the WSI method to outliers. We use the first option for the Finnish test set, as we observed that the words in the Finnish development set rarely have more than one gained sense. On the contrary, the words in the Russian development set have many gained senses, therefore, we employ \mbox{w/ WSI} for the Russian test set. For the surprise language Outlier2Cluster$_{fi}$ employs \mbox{w/o WSI} and Outlier2Cluster$^{ru}$ employs \mbox{w/ WSI}.

All of the described methods are briefly summarized in Table \ref{tab: methods}.

\section{Evaluation setup}

The first AXOLOTL-24 subtask evaluates semantic change modeling systems in three diachronic datasets in Finnish, Russian, and German~\citep{fedorova-etal-2024-axolotl}. Train and development sets are provided for the first two, but not for the last. We will now describe the datasets in more detail.
\subsection{Data sources}
The source for the Finnish dataset of the shared task is \citet{FiSrc}. The usages are divided into two groups: before 1700 and after 1700. The usages in the dataset are not complete sentences but short phrases. Some parts of the phrase can be missing and replaced with double hyphens, presumably due to OCR errors. Furthermore, the usages from both the old and the new corpus exhibit notable differences from modern Finnish. They often feature characters (such as c, z, w, and x), that are not commonly found in contemporary Finnish. It is important to highlight that the glosses provided for word senses are in modern Finnish.

Two data sources used to create the Russian dataset are \citet{RuSrcOld} processed by \citet{RuSrcOldProcessed} and \citet{RuSrcNew}. The first one was the source of old usages and glosses, and the latter provided new usages and glosses. However, the specific procedure used to map senses between these two sources was undisclosed at the time of the competition.
Some old senses are not accompanied by old usages in the Russian datasets. Consequently, our methods for the Russian datasets do not rely on the old usages. Notably, the Russian datasets lack information regarding the position of a target word within a usage or the actual word form of the target word. As a result, we incorporate the identification of the target word's position within a usage as a preprocessing step in our solution.

The shared task also includes a test dataset in a surprise language revealed only at the test phase of competition with no development or train sets. The source of this dataset is a German diachronic corpus with sense annotations \citep{schlechtweg-etal-2020-semeval,Schlechtweg2023measurement}. 

\subsection{Data Statistics}\label{sec:data stats}
\begin{figure}[t]
    \centering
    \includegraphics[width=\linewidth,height=\textheight,keepaspectratio]{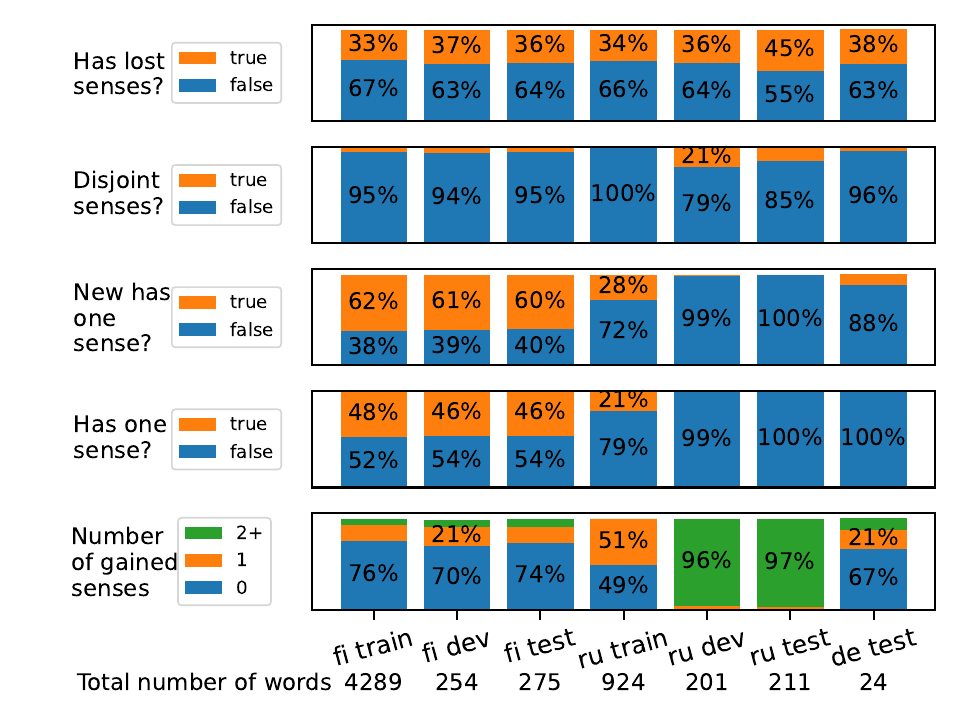}
    \caption{Proportions of target words falling into different categories in the shared task datasets. }
    \label{fig:word categories}
\end{figure}

To get insights into the data we categorize the target words within the train and the development sets based on several characteristics: \begin{itemize}
    \item Has lost senses: does the word have old senses for which there are no new usage?
    \item Number of gained senses: how many senses are there having new usages only?
    \item Disjoint senses: are the sets of senses for old and new examples disjoint?
    \item New has one sense: do all the new usages have the same meaning?
    \item Has one sense: do all the usages (both old and new) have the same meaning?
\end{itemize}
The number of target words in each category for all\footnote{This information for the test sets was not available during the competition.} the datasets of the shared task is presented on Figure~\ref{fig:word categories}. 

In the Finnish datasets, almost half of target words have only one sense and approximately 70\% of words have no gained senses. Therefore, the conservative methods that rarely discover gained senses are preferable for the Finnish datasets.

The main observation for the Russian datasets is the dramatic differences in proportions of almost all categories between the train and the development set. We can see that the statistics of the test set are similar to those of the development set. Contrary to the Finnish sets, almost all words in the Russian development set have gained senses. Therefore, methods which are prone to predict new senses rather than old ones are preferable for the Russian development set.

The German dataset is relatively small and contains ~8 times fewer words than the other test sets. We can see that it is similar to the Finnish datasets in the proportion of gained and lost senses.

\subsection{Metrics}\label{sec:metrics}

\begin{table*}[t]
\footnotesize
\begin{center}
    \begin{tabular}{||l||rrr|r|r||rrr|r|r||}
\hline

    & \multicolumn{5}{c||}{\textbf{ARI}} & \multicolumn{5}{c||}{\textbf{F1}} \\
    \textbf{Method} & \textbf{Fi} & \textbf{Ru} & \textbf{De} & \textbf{FiRu} & \textbf{AVG} & \textbf{Fi} & \textbf{Ru} & \textbf{De} & \textbf{FiRu} & \textbf{AVG} \\ 

    \hline
    \hline
    \multicolumn{11}{||l||}{ \textbf{WSD methods}}\\
    \hline

GR &0.581 &0.041 &0.386 &0.311 &0.336 &0.690 & $\diamond$\text{0.721} &0.694 &0.706 &0.702 \\
GR FiEnRu &$\diamond$\textbf{\underline{0.649}} &0.048 & $\diamond$0.521 &\textbf{0.348} & 0.406 &$\diamond$\textbf{\underline{0.756}} &$\diamond$\textbf{\underline{0.750}} &$\diamond$0.745 & $\diamond$\textbf{\underline{0.753}} & $\diamond$\textbf{\underline{0.750}} \\
GR Ru &0.568 &0.053 &0.464 &0.310 &0.361 &0.568 &$\diamond$\textbf{\underline{0.750}} &$\diamond$0.724 &0.659 &0.681 \\
GR Fi SG & $\diamond$0.638 &\textbf{0.059} &$\diamond$\textbf{\underline{0.543}} &\textbf{0.348} & \textbf{0.413} &$\diamond$0.752 &$\diamond$0.729 &$\diamond$\textbf{\underline{0.758}} &$\diamond$0.741&$\diamond$0.746 \\
    \hline
    \hline
    \multicolumn{11}{||l||}{\textbf{WSI methods}}\\
    \hline
    Agglomerative &\textbf{0.209} &$\diamond$\textbf{\underline{0.259}} &\textbf{0.316} &\textbf{0.234} &\textbf{0.261} &\textbf{0.055} &\textbf{0.152} &\textbf{0.042} &\textbf{0.104} &\textbf{0.083} \\
\hline
\hline
\multicolumn{11}{||l||}{\textbf{SCM methods}}\\
\hline
AggloM &0.581 &0 &\textbf{0.492} &0.290 &0.357 &0.674 &0 &0.695 &0.337 &0.456 \\
AggloM FiEnRu &$\diamond$0.631 &0 &0.485 & 0.315 &0.372 &$\diamond$0.731 &0 &0.639 &0.366 &0.457 \\
Cluster2Sense &0.209 &  $\diamond$\textbf{\underline{0.259}} & 0.316 &0.234 &0.261 &0.432 &0.346 &0.432 &0.389 &0.403 \\
Outlier2Cluster \scriptsize{$\begin{matrix}ru\\fi\end{matrix}$} & $\diamond$\textbf{\underline{0.649}} & $\diamond$0.247 & \scriptsize{$\begin{matrix}\textit{0.322}\\\textit{0.480}\end{matrix}$} & $\diamond$\textbf{\underline{0.448}} & \scriptsize{$\begin{matrix}\textit{0.406}\\ \diamond\textbf{\underline{\textit{0.459}}}\end{matrix}$} & $\diamond$\textbf{\underline{0.756}} & \textbf{0.645} & \scriptsize{$\begin{matrix}\textit{0.510}\\ \diamond\textbf{\textit{0.745}}\end{matrix}$} & \textbf{0.701} & \scriptsize{$\begin{matrix}\textit{0.637}\\ \diamond\textbf{\textit{0.715}}\end{matrix}$} \\

\hline
\hline
\multicolumn{11}{||l||}{\textbf{Other teams}}\\
\hline
Holotniekat &\textbf{0.596} &0.043 &0.298 &\textbf{0.319} &\textbf{0.312} &\textbf{0.655} &\textbf{0.661} &0.608 &\textbf{0.658} &\textbf{0.641} \\
TartuNLP &0.437 &0.098 &\textbf{0.396} &0.267 &0.310 &0.550 &0.640 &0.580 &0.595 &0.590 \\
IMS\_Stuttgart &0.548 &0 &0.314 &0.274 &0.287 &0.590 &0.570 &0.300 &0.580 &0.487 \\
ABDN-NLP &0.553 &0.009 &0.102 &0.281 &0.221 &\textbf{0.655} &0 &\textbf{0.638} &0.328 &0.431 \\
WooperNLP &0.428 &\textbf{0.132} &0 &0.280 &0.186 &0.503 &0.446 &0 &0.475 &0.316 \\
Baseline &0.023 &0.079 &0.022 &0.051 &0.041 &0.230 &0.260 &0.130 &0.245 &0.207 \\
\hline

\end{tabular}
\end{center}
\caption{The results on the test tests. The best result for each metric is \underline{underlined}, the best result in each group is in \textbf{bold font}. A diamond ($\diamond$) denotes those results that are worse than the best one, but the difference is practically insignificant (we consider relative differences smaller than 0.05 as practically insignificant). The official AXOLOTL-24 leaderboard is based on the average metrics across the languages having the training sets provided (the FiRu columns) and all languages (the AVG columns).}
\label{tab: results}
\end{table*}

The shared task employs two metrics to evaluate the systems, the Adjusted Rand Index (ARI) and the F1 score. 

ARI \citep{Hubert1985ComparingP} is a well-established clustering metric employed to evaluate how well new usages are clustered by a system. In the subtask, ARI is computed for all the new usages of a target word, the ground truth clusters correspond to senses. Notably, cluster labels are not taken into account by ARI. It means that old senses and gained senses are indistinguishable from each other in terms of ARI. 

The F1 score is used in the first subtask to estimate how well a system can discriminate between old senses. It is computed only for the new usages of the old senses, and not for the usages of the gained senses. The F1 score for a target word is the average of the F1 scores for all old senses. If a target word does not have any new usages with the old senses, it is arbitrarily assigned the F1 score of 1 if old senses are not predicted for any of its usages and 0 otherwise. Thus, in this edge case a system is heavily penalized when even a single usage is misclassified as one of the old senses.

All new usages of the old senses which are (incorrectly) predicted as belonging to a gained sense are considered to belong to a single auxiliary "novel" class when calculating the F1 score. The F1 score for this class is zero as it has zero precision. For this reason, even a single usage misclassified as a gained sense can dramatically affect the overall score for a target word independently of the total number of its usages.\footnote{Assume the target word has $k$ old senses. In case when only old senses are predicted: $F=\frac{F_1+...+F_k}{k}.$ If we replace one of the correct predictions of sense 1 with an incorrect prediction of a gained sense: $F'=\frac{F_{1}'+...+F_k\textbf{+0}}{\textbf{k+1}}<\frac{F_{1}+...+F_k\textbf{+0}}{\textbf{k+1}}.$ The drop in this metric is $\frac{F}{F'}>\frac{k+1}{k}$ E.g. in the case $k=1$, which is a frequent case in the Finnish AXOLOTL-24 dataset, an incorrect prediction of a gained sense for a single usage results in more than 2x decrease of the F1 score.}


\section{Results}
\subsection{Our submissions}

The number of submissions for the test sets per team was not limited in the competition. We evaluate ten models on the test sets: four WSD models (based on GlossReader, GlossReader FiEnRu, GlossReader Ru, and GlossReader Fi SG), one WSI model (Agglomerative with GlossReader representations), two AggloM models (based on GlossReader and GlossReader FiEnRu representations), Cluster2Sense, and Outlier2Cluster with different configurations for the German dataset: Outlier2Cluster$^{ru}$ and Outlier2Cluster$_{fi}$. Table \ref{tab: results} demonstrates the evaluation results. We also include the best submissions from other teams for comparison. 

\textbf{WSD and WSI.} The best results in terms of the F1 score are achieved by pure WSD methods. The F1 score is calculated only for the usages of old senses, this gives a huge advantage to WSD methods because incorrect prediction of old senses for usages of gained senses is not penalized, while the opposite reduces the F1 score severely as explained in Section~\ref{sec:metrics}.

WSD methods have notably higher ARI than Agglomerative and Cluster2Sense (both of them predict the same clusters but label them differently) for the Finnish and German datasets. On the contrary, Agglomerative and Cluster2Sense are the best-performing methods for the Russian dataset. Our explanation for this fact comes from the analysis in Section~\ref{sec:data stats}. The sets of senses of the new and the old usages in the Finnish and German datasets overlap heavily, which is beneficial for WSD methods. The overlap is much smaller for the Russian dataset, which hurts ARI of the WSD methods. Discovering gained senses is crucial for the Russian dev and test set.

\textbf{AggloM.} The AggloM method with the hyperparameter $k=0$ (never predicts gained senses) does not fall far behind pure WSD methods. The main reasons for that probably are the usage of the same underlying context encoder and prediction of only old senses. Therefore, AggloM is a viable alternative to the GlossReader models when word senses are described with usage examples instead of sense definitions.

\textbf{Outlier2Cluster.} Outlier2Cluster achieves SOTA or near-SOTA ARI\footnote{We made Outlier2Cluster$_{fi}$ submissions in the competition separately for different datasets. For this reason, it was not selected as our best submission by the competition organizers.} for Russian and Finnish, but falls behind WSD methods for German, which has no labeled data to train a dedicated NSD model. However, Outlier2Cluster can discover gained senses unlike WSD methods. Thus, we consider Outlier2Cluster to be preferable for the SCM task and suggest training the NSD model for each language of interest.\footnote{We used only small development sets with $\approx200$ target words to train novel sense detection models.}

\begin{figure}
    \centering
    \includegraphics[width=\linewidth,height=\textheight,keepaspectratio]{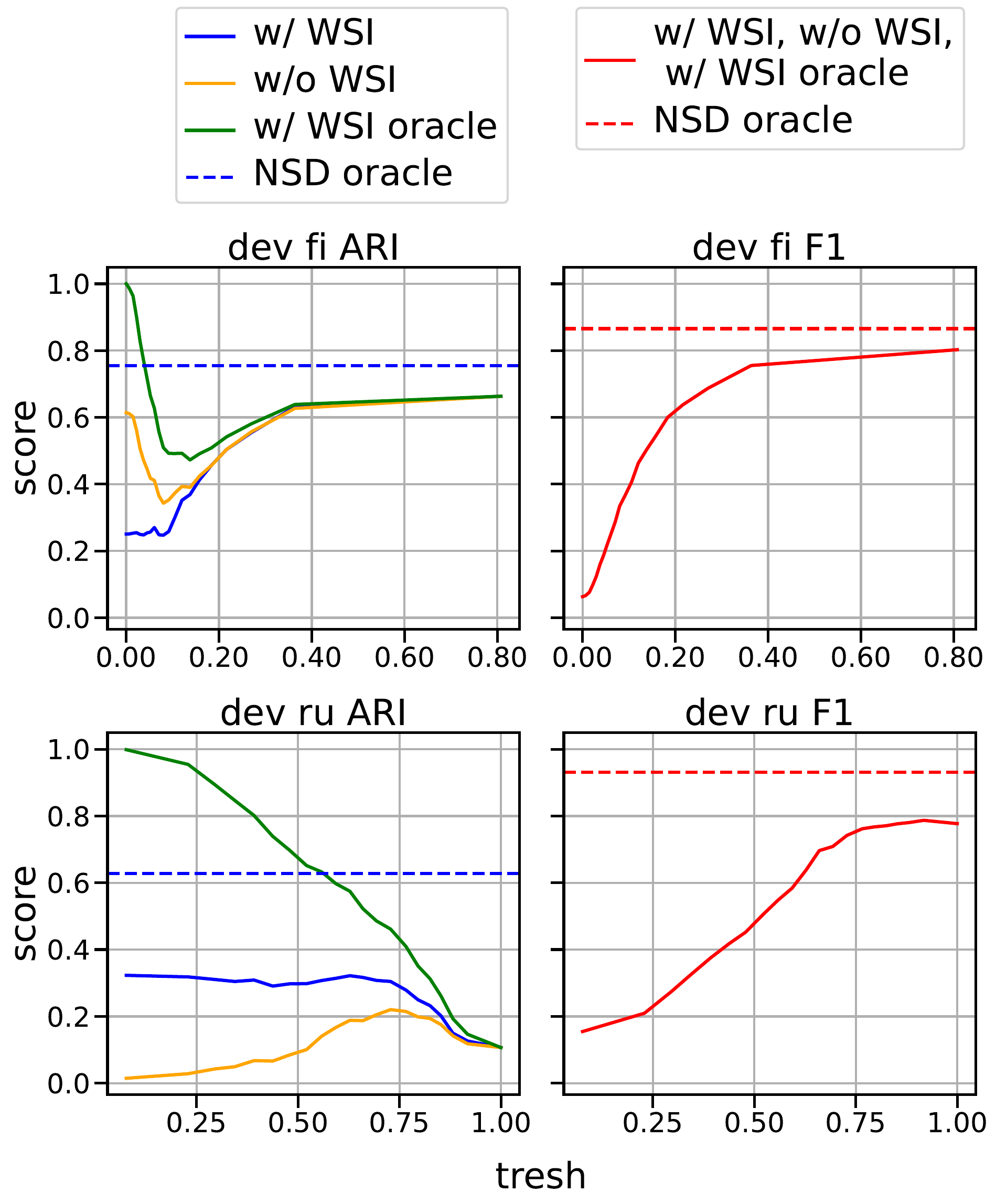}
    \caption{ARI and F1 on the development sets depending on the threshold of novel sense detector. Higher threshold means higher proportion of WSD predictions and less WSI predictions.}
    \label{fig:od thresh}
\end{figure}

The important hyperparameter of the NSD model, and consequently the Outlier2Cluster model exploiting it as a component, is the threshold dividing usages into outliers and normal usages. Figure \ref{fig:od thresh} shows the dependence of the metrics on the threshold value for the Finnish and Russian development sets. Both \mbox{w/ WSI} and \mbox{w/o WSI} versions of Outlier2cluster are included. We also compute the results of Outlier2Cluster with the WSI oracle which perfectly clusters the detected outliers according to their ground truth senses, and the NSD oracle which perfectly detects usages of gained senses. The methods we study in this Section are briefly summarized in Table \ref{tab:NSD thresh methods}.
\begin{table}[ht]
    \centering
    \footnotesize
    \begin{tabular}{|l|l l l|}
        \hline
        \textbf{Method} & \textbf{WSD} &\textbf{WSI} &\textbf{NSD} \\
        \hline
        \mbox{w/ WSI} & GR FiEnRu & Agglomer. & LogReg \\
        \mbox{w/o WSI} & GR FiEnRu & One cluster & LogReg \\
        \mbox{w/ WSI} oracle & GR FiEnRu & Oracle & LogReg \\
        NSD oracle & GR FiEnRu & Agglomer. & Oracle \\
        \hline
    \end{tabular}
    \caption{A brief summary of methods for the NSD threshold study.}
    \label{tab:NSD thresh methods}
\end{table}

We can see that the F1 score (computed only over new usages with old senses) monotonically increases with the increasing threshold, i.e. with fewer outliers detected. This again shows that trying to detect usages of gained senses and clean the old senses from them hurts the F1 score, supporting the criticism of this metric in Section~\ref{sec:metrics}.

ARI reaches a peak at the threshold of 0.65 for the Russian dataset with F1 being close to maximum as well. We therefore set the threshold at 0.65 for the Russian NSD model. This gives the SCM model that almost achieves the ARI of pure WSI predictions (threshold of 0) while having only a bit smaller F1 score compared to the best WSD model.

For the Finnish dataset, higher ARI monotonically increase with the threshold, i.e. with the proportion of predictions taken from the WSD model. This agrees with the observations from Table~\ref{tab: results} that the pure WSD models give the best ARI for Finnish. We can also see that the threshold values in the middle, where neither WSI nor WSD predictions are dominant, result in a significant decrease in ARI. It means, that our NSD model cannot be used effectively to combine the predictions for Finnish. We select a high threshold of 0.65 for the Finnish dataset, resulting in a low number of outliers. Consequently, the novel sense detector predicts less than 1\% of usages to be outliers in the Finnish test set, compared to 42\% of usages predicted as outliers for the Russian test dataset.

We can observe that according to the F1 score, the NSD oracle performs better than the pure WSD method, especially on the Russian development set. The reason lies in the words with disjointed senses. Since there are no new usages of old senses for such words, the ordinary F1 score and it is arbitrarily defined as 1 if all usages are recognized as usages of gained senses, i.e. put into new clusters, and 0 otherwise. Thus, the ideal processing of these edge cases is crucial for the F1 score, but can hardly be achieved unless the NSD oracle is employed. For other words it does not help. Considering ARI, the NSD oracle performs much better than w/WSI on the Russian dataset. It means that better NSD models may help greatly improve clustering.

According to the results of w/ WSI oracle on the Finnish development set, it is impossible to increase ARI with better WSI method without a huge drop in the F1 score. For the Russian dataset situation is the opposite. The main reason is likely the average number of gained senses per word in these datasets as described in Section~\ref{sec:data stats}. Only 7\% of words in the Finnish dataset have gained two or more senses, therefore the perfect clustering of the gained senses does not increase the results significantly compared to merging all gained senses into a single cluster. On the contrary, 97\% of the word in the Russian have two or more gained senses, making WSI necessary.

\section{Conclusion}
We have proposed three new methods that solve the SCM task. Our solution achieves SOTA results among all participants of the first subtask of the AXOLOTL-24 shared task. Additional experiments propose directions of further improvement of the developed models, NSD being potentially the most promising one.

\section{Limitations}
While our methods can in theory be applied to any SCM dataset, we acknowledge that they may be overspecified for the first subtask of AXOLOTL-24. Notably, we extensively use the train sets provided for the competition in Finnish and Russian to train the embedding model and to optimize the hyperparameters. While we also evaluate on the German dataset in a zero-shot fashion, the results may be unreliable due to relatively small size of the dataset. 

Semantic change modeling may be of particular interest in studies of older time periods, where the language is quite different from its modern state. The underlying model, GlossReader, is a finetuned version of XLM-R, which was not specifically designed to handle old languages. In this case dataset-specific finetuning of the base GlossReader may become even more relevant.

\section{Acknowledgements}
Nikolay Arefyev has received funding from the European Union’s Horizon Europe research and innovation program under Grant agreement No 101070350 (HPLT).

\bibliography{acl_latex}

\appendix
\section{Ablation study of the NSD model}
\label{appendix: novel sense detection}

\begin{table*}[!ht]
\footnotesize
\centering
\begin{tabular}{||l||cc||cc||}
\hline
 & \multicolumn{2}{c||}{\textbf{dev fi AP}} & \multicolumn{2}{c||}{\textbf{dev ru AP}} \\
\textbf{Model} & \multicolumn{1}{c|}{\textbf{GR}} &  \textbf{GR FiEnRu} & \multicolumn{1}{c|}{\textbf{GR}} & \textbf{GR FiEnRu} \\ 
\hline
\hline
\multicolumn{5}{||l||}{\textbf{single features}}\\

\hline
cosine &\multicolumn{1}{c|}{0.106}  & 0.110  &\multicolumn{1}{c|}{0.685}  & 0.695 \\
euclid. &\multicolumn{1}{c|}{0.106}  & 0.110 & \multicolumn{1}{c|}{0.684} & 0.694 \\
l2/euclid. & \multicolumn{1}{c|}{0.106} & 0.110 &\multicolumn{1}{c|}{0.685}  & 0.695 \\
manh. & \multicolumn{1}{c|}{0.106} & 0.113 & \multicolumn{1}{c|}{0.685} & 0.690 \\
l1/manh. & \multicolumn{1}{c|}{\textbf{0.154}} & \textbf{0.242} &\multicolumn{1}{c|}{\textbf{0.816}}  & \textbf{0.822} \\

\hline
\hline
\multicolumn{5}{||l||}{\textbf{\textbf{full classifiers}}}\\
\hline

classifier w/ extra & \multicolumn{2}{c||}{\textbf{\underline{0.378}}} & \multicolumn{2}{c||}{\textbf{\underline{0.840}}} \\
classifier w/o extra & \multicolumn{2}{c||}{0.305} & \multicolumn{2}{c||}{0.833} \\ 

\hline
\hline
\multicolumn{5}{||l||}{\textbf{\textbf{best pairs of features w/o extra features}}}\\
\hline

 l1/manh. + euclid. & \multicolumn{1}{c|}{0.194} & \multicolumn{1}{c||}{\textbf{{0.284}}} & \multicolumn{1}{c|}{0.818} & \multicolumn{1}{c||}{\textbf{0.823}}  \\
 l1/manh. + l2/euclid. & \multicolumn{1}{c|}{\textbf{{0.195}}} & \multicolumn{1}{c||}{\textbf{0.284}}& \multicolumn{1}{c|}{0.818} & \multicolumn{1}{c||}{\textbf{0.823}}  \\
 l1/manh. + manh. & \multicolumn{1}{c|}{0.192} & \multicolumn{1}{c||}{0.277}& \multicolumn{1}{c|}{\textbf{{0.819}}} & \multicolumn{1}{c||}{\textbf{{0.823}}}  \\

 \hline
 \hline
\multicolumn{5}{||l||}{\textbf{\textbf{best pairs of features w/ extra features}}}\\
\hline

 l1/manh. + \#old usages & \multicolumn{1}{c|}{0.190} & \multicolumn{1}{c||}{\textbf{{0.291}}} & \multicolumn{1}{c|}{0.820} & \multicolumn{1}{c||}{0.827}  \\
 l1/manh. + \#new usages & \multicolumn{1}{c|}{0.153} & \multicolumn{1}{c||}{0.249}& \multicolumn{1}{c|}{\textbf{{0.821}}} & \multicolumn{1}{c||}{\textbf{{0.829}}}  \\
 \#new usages + \#old senses & \multicolumn{1}{c|}{\textbf{{{0.266}}}} & \multicolumn{1}{c||}{0.266} & \multicolumn{1}{c|}{0.643} & \multicolumn{1}{c||}{0.643}  \\
\hline
\end{tabular}
\caption{Average precision of novel sense detection models on the dev sets. Except for the block with full classifiers, models use distance-based features either from GlossReader or GlossReader FiEnRu. The best results in each group are in \textbf{bold font}. The overall best results are \underline{underlined}.}
\label{tab: OD results}
\end{table*}

\begin{figure}[ht]
       \centering
       \includegraphics[width=\linewidth,height=\textheight,keepaspectratio]{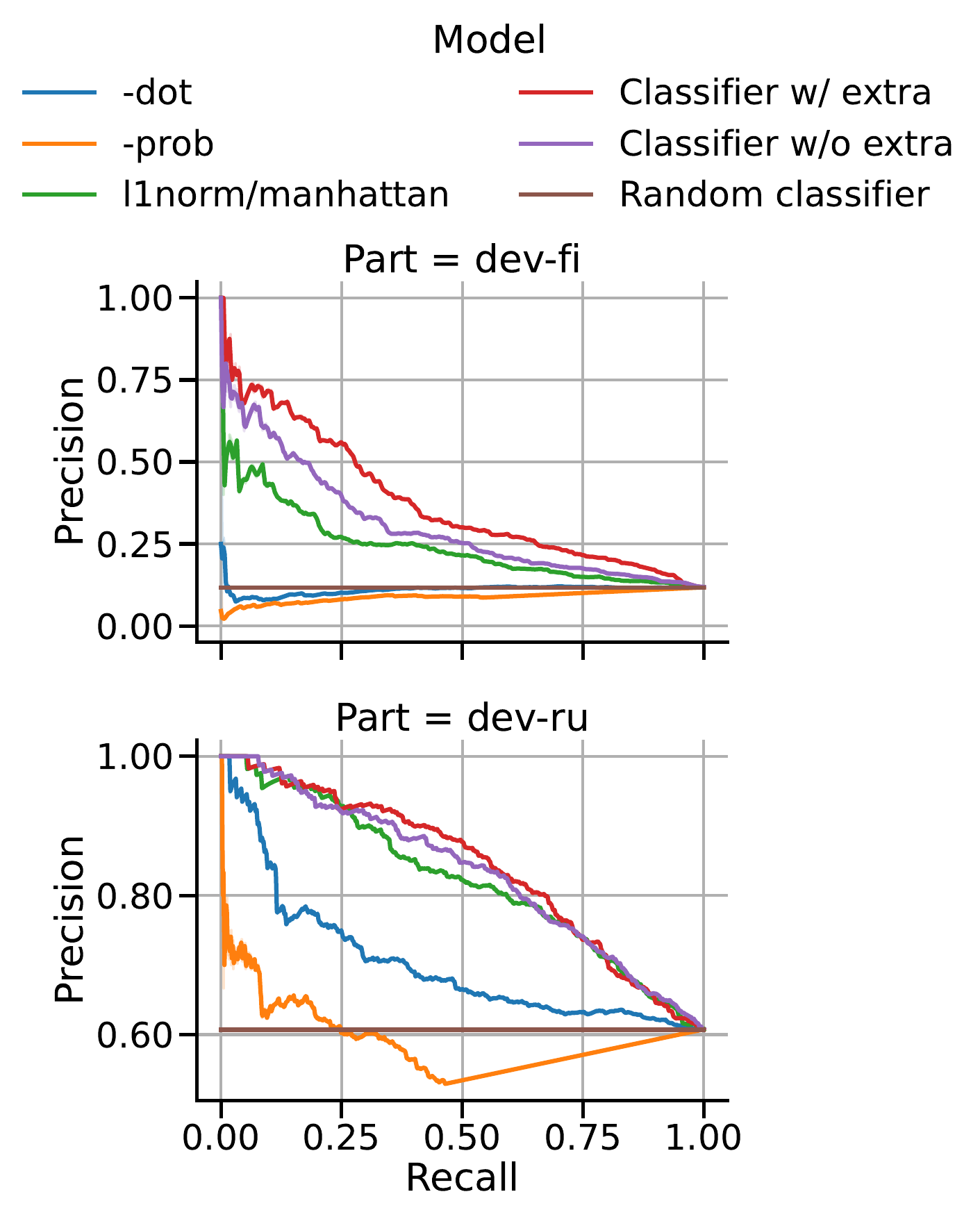}
       \caption{Precision-recall curves of novel sense detection models. Non classifier models are distances between usages and chosen glosses from GlossReader FiEnRu. Classifier w/ extra stands for classifier trained on distance-based and non distance-based features introduced in sub subsection \ref{sec: outlier2cluster}. Classifier w/o extra stands for classifier trained only on distance-based features.}
       \label{fig:od pr curve}
   \end{figure}

In this section we provide an ablation study of our NSD model. In order to get insights about the importance of the chosen features, we compare our trained classifiers and several pure similarity measures such as the predicted probability, the dot product and the distance from a usage to the gloss selected by GlossReader FiEnRu (Figure \ref{fig:od pr curve}). It turns out that the probabilities and dot products are far behind the classifiers, and on the Finnish dev set they even perform no better than a random classifier. The manhattan distance with l1 normalization is a bit worse than the trained classifiers. The extra features consistently help on the Finnish dev set, but are almost useless on the Russian dev set.

In Table \ref{tab: OD results} we compare different NSD models using the average precision on the dev sets.
To understand which quality can be achieved using the minimal number of features, we evaluate all single distance-based features. Furthermore, we train classifiers on all possible pairs of features, where each pair contains distances only from the same GlossReader. Also we compare classifiers with or without extra features.

We observe that the manhattan distance with l1 normalization, which is the best single feature, works poorly on the Finnish dataset, especially for the embeddings from GlossReader that was not fine-tuned on the Finnish train set. However, on the Russian dev set it closely follows the best classifier. As for the classifiers, we found that including non-distance features is important for Finnish. What is more interesting, when using the original GlossReader model among all pairs of features the best one does not contain embedding-based features at all, only the number of old senses and the number of new usages. This signals that for the Finish dataset GlossReader provides poor embeddings without fine-tuning.
\end{document}